\documentclass{article}
\PassOptionsToPackage{numbers, compress}{natbib}
\usepackage[final,sglblindworkshop]{neurips_2025}

\workshoptitle{NeurIPS 2025 Workshop on Efficient Reasoning}
\usepackage[most]{tcolorbox}
\usepackage[svgnames]{xcolor} % For defining colors using HTML hex codes
\usepackage{graphicx}         % For including figures
\usepackage{subfigure}        % For creating subfigures
\usepackage{placeins}         % For using \FloatBarrier
\usepackage[utf8]{inputenc} % allow utf-8 input
\usepackage[T1]{fontenc}    % use 8-bit T1 fonts
\usepackage{hyperref}       % hyperlinks
\usepackage{url}            % simple URL typesetting
\usepackage{booktabs}       % professional-quality tables
\usepackage{amsfonts}       % blackboard math symbols
\usepackage{nicefrac}       % compact symbols for 1/2, etc.
\usepackage{microtype}      % microtypography
\usepackage{xcolor}         % colors
\usepackage{graphicx}
\usepackage{subfigure}
\usepackage{booktabs} % for professional tables
\usepackage{multirow}
\usepackage[table]{xcolor}
\usepackage{multirow}
\usepackage{array}
\usepackage{tabularx} % For auto-sizing tables
\usepackage{multirow} % You already have this
\usepackage[table]{xcolor} % for row striping

\newcolumntype{C}{>{\centering\arraybackslash}X} % Defines a new centered 'X' column type
\usepackage{amsmath}
\usepackage{amssymb}
\usepackage{mathtools}
\usepackage{amsthm}
\usepackage{graphicx}
\usepackage{subcaption}   
\usepackage{placeins}     
\usepackage{booktabs} 
\newcommand{\changed}[1] {{ #1}}
\newcommand{\changedb}[1] {{ #1}}

\definecolor{lightblue}{RGB}{248,249,255}
\definecolor{lightorange}{RGB}{255,248,240}
\definecolor{lightgreen}{RGB}{240,255,248}

\newcommand{\mytodo}[1]{}

\theoremstyle{definition}

\theoremstyle{remark}

\title{The Virtues of Brevity: Avoid Overthinking in Parallel Test-Time Reasoning}

\author{
  Raul Cavalcante Dinardi\textsuperscript{1}\hspace{0.65em}%
  Bruno Yamamoto\textsuperscript{2}\hspace{0.65em}%
  Anna Helena Reali Costa\textsuperscript{2}\hspace{0.65em}%
  Artur Jordao\textsuperscript{2} \\
  {\normalfont \textsuperscript{1}\,Instituto de Matemática, Estatística e Ciência da Computação, Universidade de São Paulo} \\
  {\normalfont \textsuperscript{2}\,Escola Politécnica, Universidade de São Paulo} \\
  {\normalfont \texttt{raulcdinardi@usp.br}}
}

\begin{document}

\maketitle
\begin{abstract}
Reasoning models represent a significant advance in LLM capabilities, particularly for complex reasoning tasks such as mathematics and coding.
Previous studies confirm that parallel test-time compute—sampling multiple solutions and selecting the best one—can further enhance the predictive performance of LLMs.
However, strategies in this area often require complex scoring, thus increasing computational cost and complexity.
In this work, we demonstrate that the simple and counterintuitive heuristic of selecting the shortest solution is highly effective.
We posit that the observed effectiveness stems from models operating in two distinct regimes: a concise, confident \emph{conventional regime} and a verbose \emph{overthinking regime} characterized by uncertainty, and we show evidence of a critical point where the overthinking regime begins to be significant. By selecting the shortest answer, the heuristic preferentially samples from the conventional regime. We confirm that this approach is competitive with more complex methods such as self-consistency across two challenging benchmarks while significantly reducing computational overhead. The shortest-answer heuristic provides a Pareto improvement over self-consistency and applies even to tasks where output equality is not well defined.

\end{abstract}
\section{Introduction}
Large Language Models demonstrate impressive single-shot performance on a wide range of tasks~\cite{DBLP:conf/nips/BrownMRSKDNSSAA20,Bengio:2025,Maslej:2025}. Nevertheless, for highly valuable problems, one can further extend their capabilities and reliability by using more compute~\cite{li2025a,DBLP:conf/iclr/Snell0XK25}.
One method for increasing performance by scaling compute at test time is Best-of-N selection, which samples $N$ solutions from the same model for each problem, and employs a heuristic to pick the most likely to be correct. For example, self-consistency uses the heuristic of consistency~\cite{DBLP:conf/iclr/0002WSLCNCZ23}, among $N$ solutions the model generates given a prompt, it picks the answer that appears most frequently.
Still, Best-of-N heuristics suffer from downsides, some, like self-consistency, are inflexible towards non-comparable outputs, while others require dedicated reward models~\cite{DBLP:conf/iclr/ZhangHBKKA25}. Another method for test-time scaling led to a new paradigm: the reasoning model.
The training in this family of models enables them to emergently learn to use a long chain-of-thought (CoT) fruitfully~\cite{o1}, greatly improving accuracy, especially on cognitively demanding tasks such as mathematics and coding.

\begin{figure}[!t]
    \centering
    \begin{minipage}[c]{0.55\textwidth}
        \includegraphics[width=\linewidth]{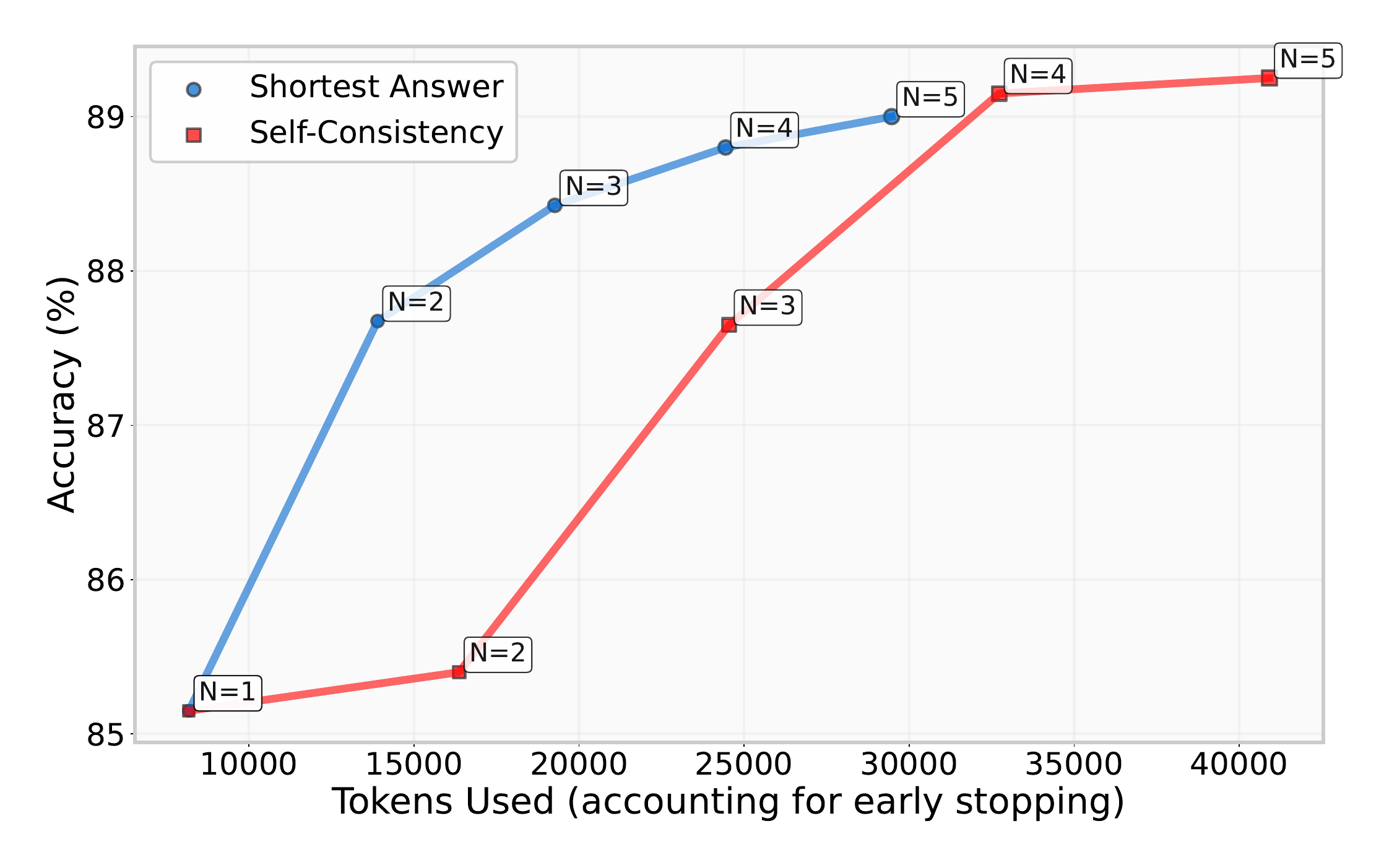}
    \end{minipage}
    \hfill
    \begin{minipage}[c]{0.4\textwidth}
        \captionof{figure}{Pareto curve of accuracy against token usage for DeepSeek-R1, comparing efficacy of self-consistency and picking the shortest solution on 400 AIME questions (Top-left is optimal). Picking the shortest solution is more token-efficient due to early stopping; once the first solution completes, we terminate all others, saving tokens. See Appendix~\ref{app:pareto_curves} for other models.}
        \label{fig:pareto}
    \end{minipage}
\end{figure}

However, recent literature shows that these reasoning models suffer from \emph{overthinking}, a phenomenon in which they generate more tokens than necessary, thereby wasting compute~\cite{DBLP:overthinking}. Prior work focuses on overthinking token waste after a model should have been confident enough to stop generating a response but continues generating on trivial problems. We demonstrate the other side of this regime: the model, when unconfident, still displays overthinking, this causes a systematic skew in solution length between correct and incorrect generations. We exploit the resulting skew in solution length as a method for parallel test-time compute scaling, picking the shortest solution from $N$ completions for each problem.

Previous efforts attribute overthinking to a bias in the training of reasoning models~\cite{liu2025understanding}, which use Reinforcement Learning with verifiable rewards. This bias is rooted in the usual algorithms for training such models: GRPO~\cite{grpo}, GSPO~\cite{gspo}, and most PPO~\cite{ppo} implementations, whose reward functions have a term that normalizes rewards by solution length. This means a model can learn the following strategy to minimize negative reward per token in training: if the model can reliably distinguish solutions that will lead to correct responses from those that will not, it can exploit this regularization. When its estimate of a solution's correctness is low, it mitigates the penalty by continuing the generation with frivolous reasoning, thus diluting the negative reward over a greater number of tokens. Other findings support this argument, reasoning models have a better ability to accurately assess the correctness of their own responses than non-reasoning LLMs~\cite{confidence_LRM}.

We build on this theoretical foundation by empirically demonstrating that reasoning models operate in two regimes, a \emph{conventional regime} characterized by shorter solutions, with both textual uncertainty and textual similarity between solutions correlating with solution length, and an \emph{overthinking regime} characterized by longer solutions where the trend in embedding distance and uncertainty breaks. Through textual analysis of CoTs, we show the critical point in token usage where the overthinking regime starts to meaningfully take over, after the peak of the token usage distribution for the conventional-regime solutions, given model and problem set we sample from, where overthinking begins to be noticeable. We observe trend breaks congruent with the two-regime hypothesis at the critical point of each model.

Based on previous observations, we explain why the shortest solution exhibits high performance across a difficult set of benchmarks in mathematics and coding. In particular, it offers discriminative power, commensurate with self-consistency at a lower cost and better theoretical end-to-end latency. Furthermore, it applies to a broader set of tasks where outputs are not directly comparable, representing a Pareto improvement over self-consistency. 
This efficiency gain stems from the early-stopping strategy in the parallel case: once a solution completes, we discard all other candidates whose costs already exceed its value. Assuming synchronous token generation, this guarantees that the terminated solutions would correspond to worse (i.e., longer) solutions.

Experiments on challenging benchmarks show that our simple short-solution heuristic matches or even surpasses advanced test-time compute strategies such as self-consistency, while being more computationally efficient. For example, on DeepSeek-R1~\cite{DBLP:journals/corr/abs-2501-12948}, our heuristic matches self-consistency and exceeds the single-solution accuracy by four percentage points (Table~\ref{tab:m3_benchmarks}). We also observe the same behavior on Grok-3-mini~\cite{grok3} and Qwen3-32B~\cite{qwen3techreport}.

\begin{table*}[!ht]
\centering
\small
\renewcommand{\arraystretch}{1.0}  % Reduced from 1.05
\setlength{\aboverulesep}{0pt}     % Explicitly set to 0
\setlength{\belowrulesep}{0pt}     % Explicitly set to 0
\arrayrulecolor{gray!50}

\begin{tabularx}{\textwidth}{>{\centering\arraybackslash}p{3.2cm} l C C}
\toprule
\textbf{Model} & \textbf{Method} & \textbf{AIME} & \textbf{LiveCodeBench} \\
\midrule

\multirow{4}{*}{\textbf{DeepSeek-R1}} 
    & \cellcolor{gray!8}Individual Attempts & \cellcolor{gray!8}85.0\% & \cellcolor{gray!8}76.5\% \\
    & \textbf{Shortest Solution}      & 89.0\% & 79.2\% \\
    & \cellcolor{gray!8}Self-Consistency    & \cellcolor{gray!8}89.2\% & \cellcolor{gray!8}\textcolor{red}{*}       \\
    & Longest Solution            & 78.2\% & 76.5\% \\
\midrule

\multirow{4}{*}{\textbf{Grok-3-mini}} 
    & \cellcolor{gray!8}Individual Attempts & \cellcolor{gray!8}81.0\% & \cellcolor{gray!8}69.5\% \\
    & \textbf{Shortest Solution}      & 85.2\% & 69.2\% \\
    & \cellcolor{gray!8}Self-Consistency    & \cellcolor{gray!8}86.2\% & \cellcolor{gray!8}\textcolor{red}{*}       \\
    & Longest Solution            & 74.9\% & 66.8\% \\
\midrule
\multirow{4}{*}{\textbf{Qwen3-32B}} 
    & \cellcolor{gray!8}Individual Attempts & \cellcolor{gray!8}89.5\% & \cellcolor{gray!8}78.6\% \\
    & \textbf{Shortest Solution}      & 92.5\% & 79.5\% \\
    & \cellcolor{gray!8}Self-Consistency    & \cellcolor{gray!8}93.0\% & \cellcolor{gray!8}\textcolor{red}{*}       \\
    & Longest Solution            & 85.5\% & 76.8\% \\
\bottomrule
\end{tabularx}
\caption{Best-of-N heuristic comparison, with $N=5$, on AIME and LiveCodeBench benchmarks. The symbol \textcolor{red}{*} means self-consistency is not applicable for the task due to non-comparable answers.}
\label{tab:m3_benchmarks}
\end{table*}
\section{Experiments}

\textbf{Experimental Setup.}
We conduct experiments on three models: DeepSeek-R1~\cite{DBLP:journals/corr/abs-2501-12948}, Grok-3-mini~\cite{grok3}, and Qwen3-32B~\cite{qwen3techreport} using a random subset of 400 questions from the AIME math competition~\cite{aime_1983_2024} and LiveCodeBench v5~\cite{DBLP:conf/LiveCodeBench}.
We run each model with $\texttt{temperature}=1$. Throughout the experiments, we sample five solutions per problem and use minimal prompts for each benchmark to elicit the model to respond in an extractable format (we describe the full prompts in \autoref{app:prompts}). We compute textual embeddings using a pre-trained sentence transformer, specifically \texttt{all-MiniLM-L6-v2} ~\cite{miniLM}. %\cite{reimers-2019-sentence-bert}.

\textbf{Quantitative Results.}
Table~\ref{tab:m3_benchmarks} summarizes the results. From this table, the shortest length heuristic is able to match the performance of self-consistency~\cite{DBLP:conf/iclr/0002WSLCNCZ23}. Also, selecting the longest solution rather than the shortest yields lower accuracy than individual attempts, aligning with our overthinking hypothesis. By analyzing the relationship between accuracy $\times$ number of tokens (Figure~\ref{fig:pareto}), we observe that it grows sublinearly.

In this context, it is particularly informative to examine the behavior at the lowest $N$ values, where the rate of improvement is highest.
By the nature of the method, we can discriminate between a pair of solutions per problem ($N=2$), and observe great improvement, in contrast to self-consistency where the method necessitates at least 3 solutions so that there can be a consensus among the solutions ($N\geq 3$). Thus, shortest solution selection can be more practically beneficial in cost-constrained scenarios.

\textbf{Uncertainty in Reasoning Models.} \changedb{To understand why shorter solutions are often correct, we analyze a key component of reasoning—uncertainty—and its relation to overthinking.} For this purpose, we perform linguistic analysis on the frequency of uncertainty markers (e.g. \emph{maybe}, \emph{alternatively}, \emph{but}, \emph{wait}, \emph{perhaps}; ~\autoref{app:uncertainty_markers} \changed{provides the full list of markers}) in the chain-of-thought text of each solution across multiple models.
%\hyperref[tab:uncertainty_comparison]{Table~\ref*{tab:uncertainty_comparison}} 
%Table~\ref{tab:uncertainty_comparison} demonstrates that, across all models, longer responses exhibit more uncertainty markers than shorter ones,
Table~\ref{tab:uncertainty_comparison} shows that, across all models, longer responses exhibit \changedb{a higher density of} uncertainty markers than shorter ones, even when both answer the same problem correctly. This supports our hypothesis that models tend to continue generating when uncertain.
\begin{table}[!ht]

	\centering
	\small
	\renewcommand{\arraystretch}{1.0}
	\setlength{\aboverulesep}{0pt}
	\setlength{\belowrulesep}{0pt}
	\arrayrulecolor{gray!50}
	
\begin{tabular}{>{\centering\arraybackslash}p{3cm} >{\centering\arraybackslash}p{2.5cm} >{\centering\arraybackslash}p{2.5cm}}
  \toprule
  \textbf{Model} & \multicolumn{2}{c}{\textbf{Cases where longest is more uncertain}} \\
  \cmidrule{2-3}
                  & AIME & LiveCodeBench \\
  \midrule
  DeepSeek-R1  & \cellcolor{gray!8}67.0\% & \cellcolor{gray!8}67.5\% \\
  Grok-3-mini & 67.4\% & 63.7\% \\
  Qwen3-32B & \cellcolor{gray!8}58.2\% & \cellcolor{gray!8}65.8\% \\
  \bottomrule
\end{tabular}
\caption{Percentage of cases where the longer of two correct solutions to the same problem exhibits higher uncertainty density (normalized per 100 words).}
\label{tab:uncertainty_comparison}
\end{table}

\begin{figure}[!t]
\centering
\subfigure[Token usage distribution for DeepSeek-R1 longest and shortest solutions to each problem, showing that the mode of the distribution stays the same, the shortest solution distribution is different as it diminishes the overthinking long-tail.\label{fig:token_dist_deepseek}]{%
\includegraphics[width=0.3\columnwidth]{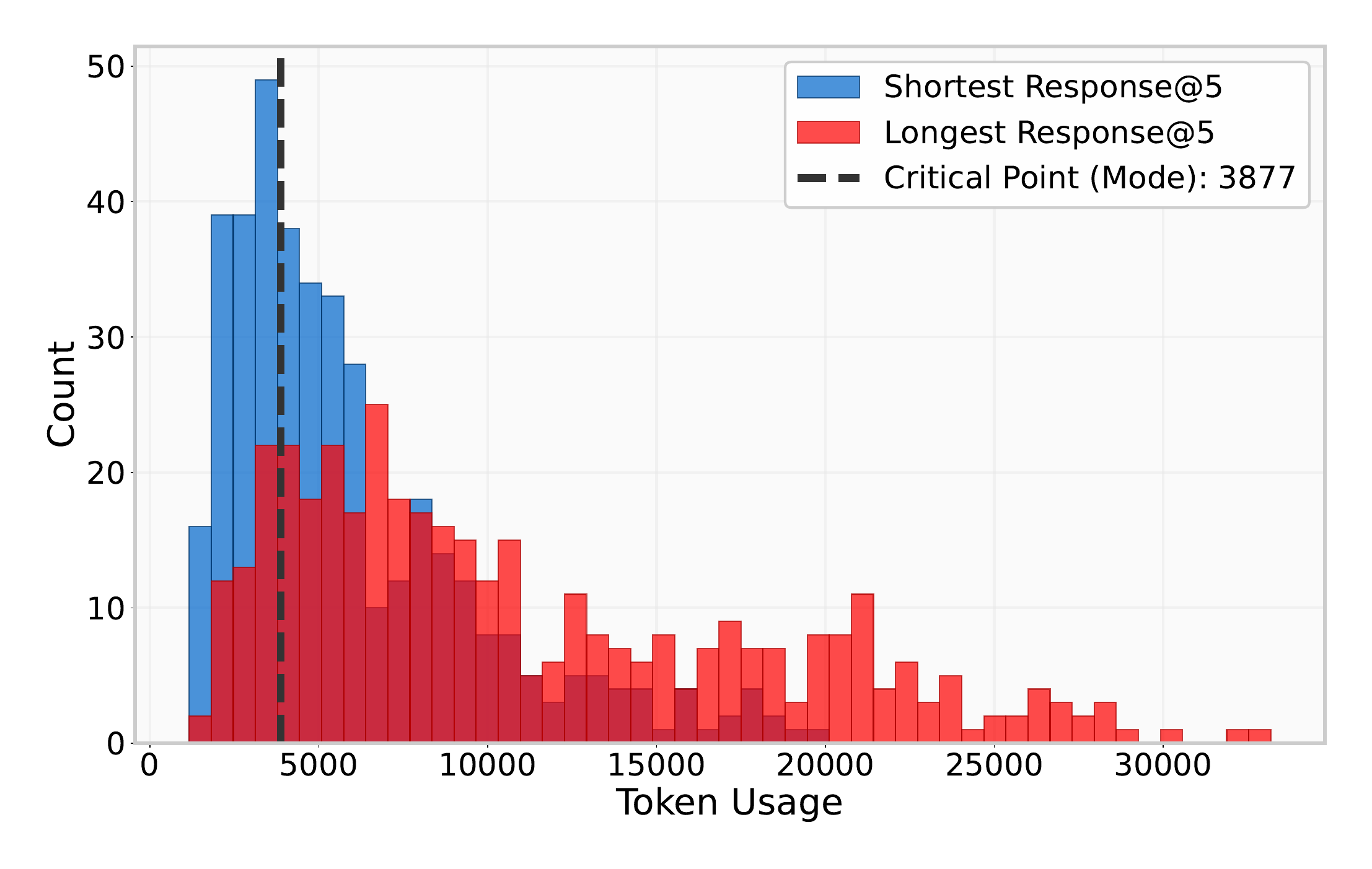}}
\hfill
\subfigure[Before/after critical point analysis for frequency of uncertainty markers. Points below the critical point (blue) show different uncertainty patterns than those above (red), with distinct regression slopes.\label{fig:before_after_deepseek}]{%
\includegraphics[width=0.3\columnwidth]{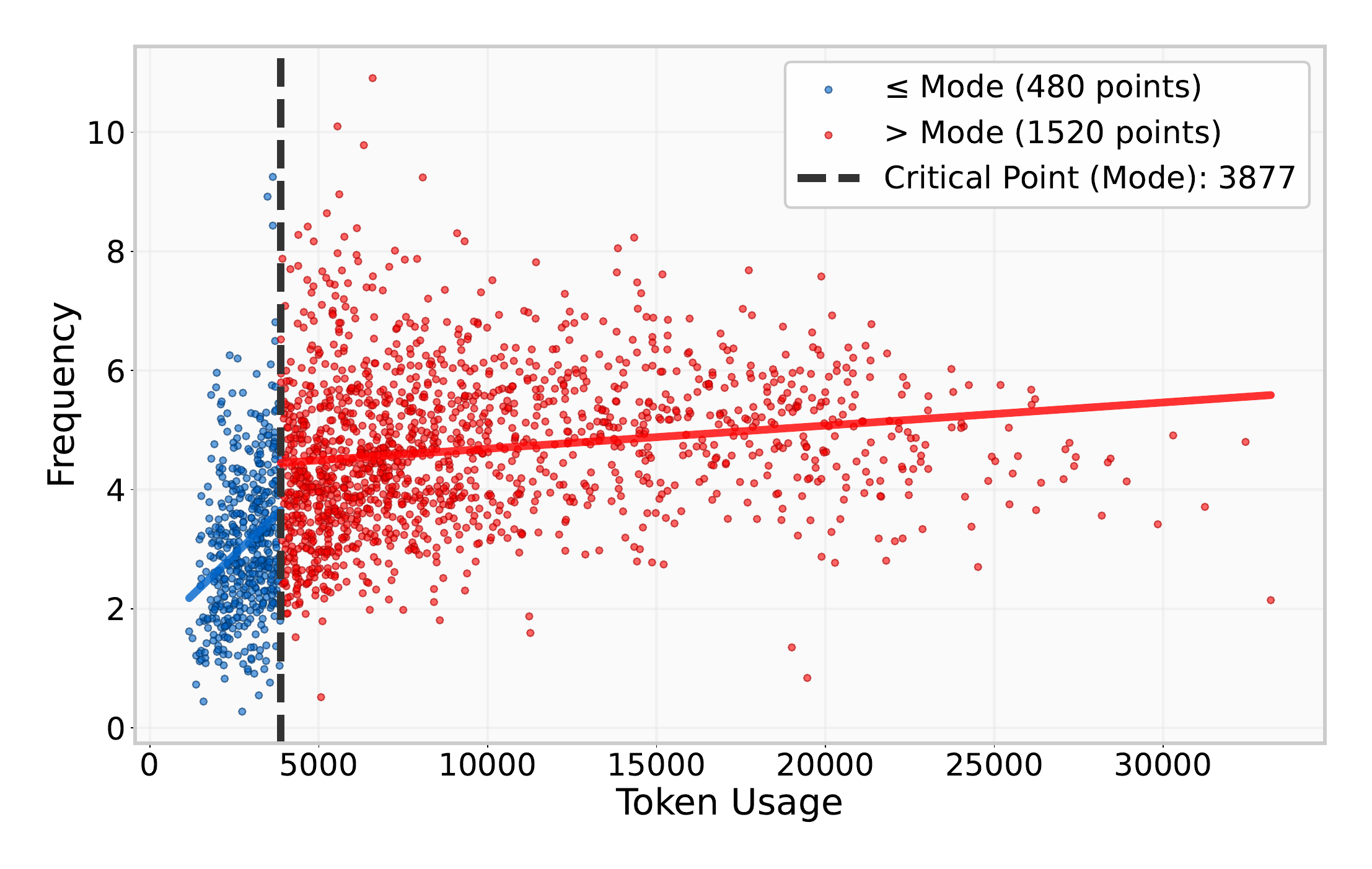}}
\hfill
\subfigure[Average pairwise cosine distance between embeddings of 500-word CoT chunks from parallel solutions to the same problems. The distance rises until the critical point before plateauing.\label{fig:embedding_divergence}]{
\includegraphics[width=0.34\columnwidth]{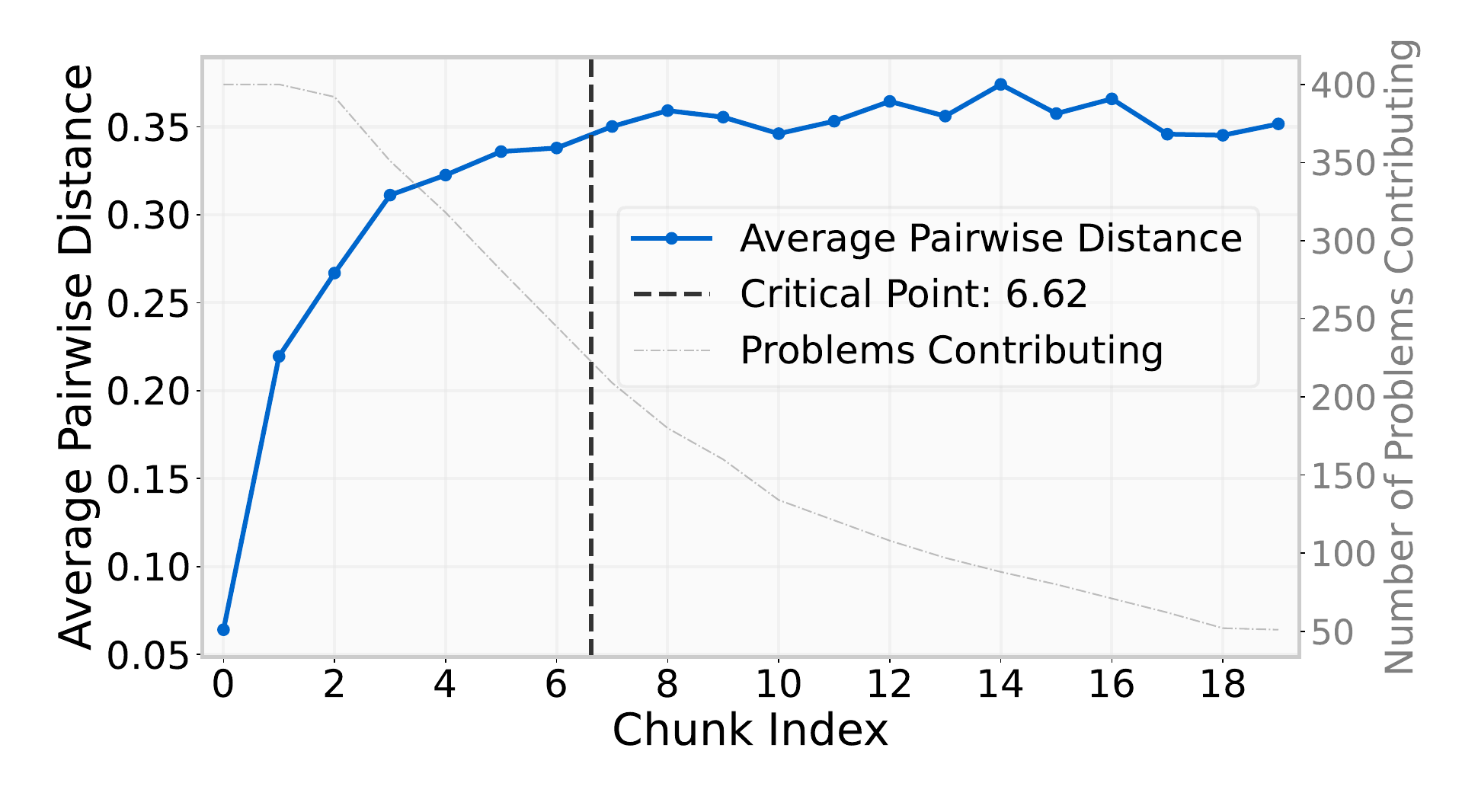}}
\label{fig:combined}

\caption{Analysis of different trend breaks for DeepSeek-R1 on the AIME benchmark before and after the critical point, which indicates the separation between the conventional and overthinking regimes. See Appendix~\ref{app:critical_point} for other models.}
\end{figure}

To further corroborate our two-regime hypothesis, we analyze solutions before and after a critical threshold where we expect the overthinking regime to start to take over. We define this threshold as the mode of the overall token usage distribution since this mode serves as a reasonable proxy for the peak of the token-usage distribution for conventional-regime solutions. This point marks the juncture where the proportion of overthinking solutions relative to conventional ones starts to rise, identifying where the overthinking regime starts to become dominant.
Therefore, we hypothesize that the modal length marks the onset of overthinking, characterized by extraneous generation that artificially lengthens solutions. To test this, we examine the frequency of uncertainty markers and observe a distinctive trend break around the hypothesized critical point, indicating a genuine regime change (Figure~\ref{fig:before_after_deepseek}). Before the critical point, solution length is highly positively correlated with the expression of uncertainty markers; however, after this point, the trend breaks.

\textbf{Intra-problem solution differences.}
We further analyze our results by looking at the intra-problem differences between pairs of solutions to the same problem.
First, we show that the token-usage distribution for the shortest solution among the $N$ we generate for each problem is less skewed than the longest solution, as suggests the distribution shape in Figure~\ref{fig:token_dist_deepseek}, but the mode is the same, providing further evidence that the heuristic works by avoiding the overthinking tail.

We also analyze the divergence in reasoning paths, we compare every pair of solutions we generate for each problem. We partition each solution's CoT text into 500-word chunks and compute the embedding vector of each chunk using \texttt{all-MiniLM-L6-v2}. Then, we compare each chunk to the chunk in the same position in the other solution (i.e., 500-word chunk which starts and ends in the same absolute position) and calculate the cosine distance between their embeddings. Figure~\ref{fig:embedding_divergence} shows that the cosine distance between pairs of solution chunks rises until the critical points and remains constant afterwards, strengthening our two regime hypothesis.

\section{Conclusions}
In this work, we demonstrate a cost-effective heuristic for parallel test-time compute scaling: generating multiple solutions and selecting the shortest one. Through early-stopping, it achieves a Pareto improvement in the token-usage-to-performance curve over the popular self-consistency.
We attribute the efficacy of this heuristic to its ability to avoid the overthinking regime. Our analysis shows that this phenomenon is not limited to trivial tasks but extends to complex problems, where it significantly decreases the efficiency of token-usage. By opting for brevity, the heuristic effectively leverages the model's self-assessment of correctness, explaining its counterintuitive and surprising success.

\renewcommand{\acksection}{\section*{Acknowledgments}}
\begin{ack}
This study was financed in part by the Coordenação de Aperfeiçoamento de Pessoal de Nível Superior -- Brasil (CAPES)
-- Finance Code 001. Artur Jordao Lima Correia would like to thank Edital Programa de Apoio a Novos Docentes 2023.
Processo USP n\textsuperscript{o}: 22.1.09345.01.2. Anna H. Reali Costa would like to thank grant \#312360/2023-1 CNPq.
\end{ack}
\clearpage
\bibliography{refs}

%%%%%%%%%%%%%%%%%%%%%%%%%%%%%%%%%%%%%%%%%%%%%%%%%%%%%%%%%%%%
\newpage
\appendix

\section{Technical Appendices and Supplementary Material}
This section provides analysis results across all models and benchmarks we test and additional details. We use the same setup as described in the main text.

% NOTE: Ensure you have \usepackage{placeins} in your preamble for \FloatBarrier to work.
\FloatBarrier 
\subsection{Pareto Performance Curves}
\label{app:pareto_curves}
\begin{figure}[htbp]
\centering
\subfigure[Pareto curve for Grok-3-mini.]{
\includegraphics[width=0.48\textwidth]{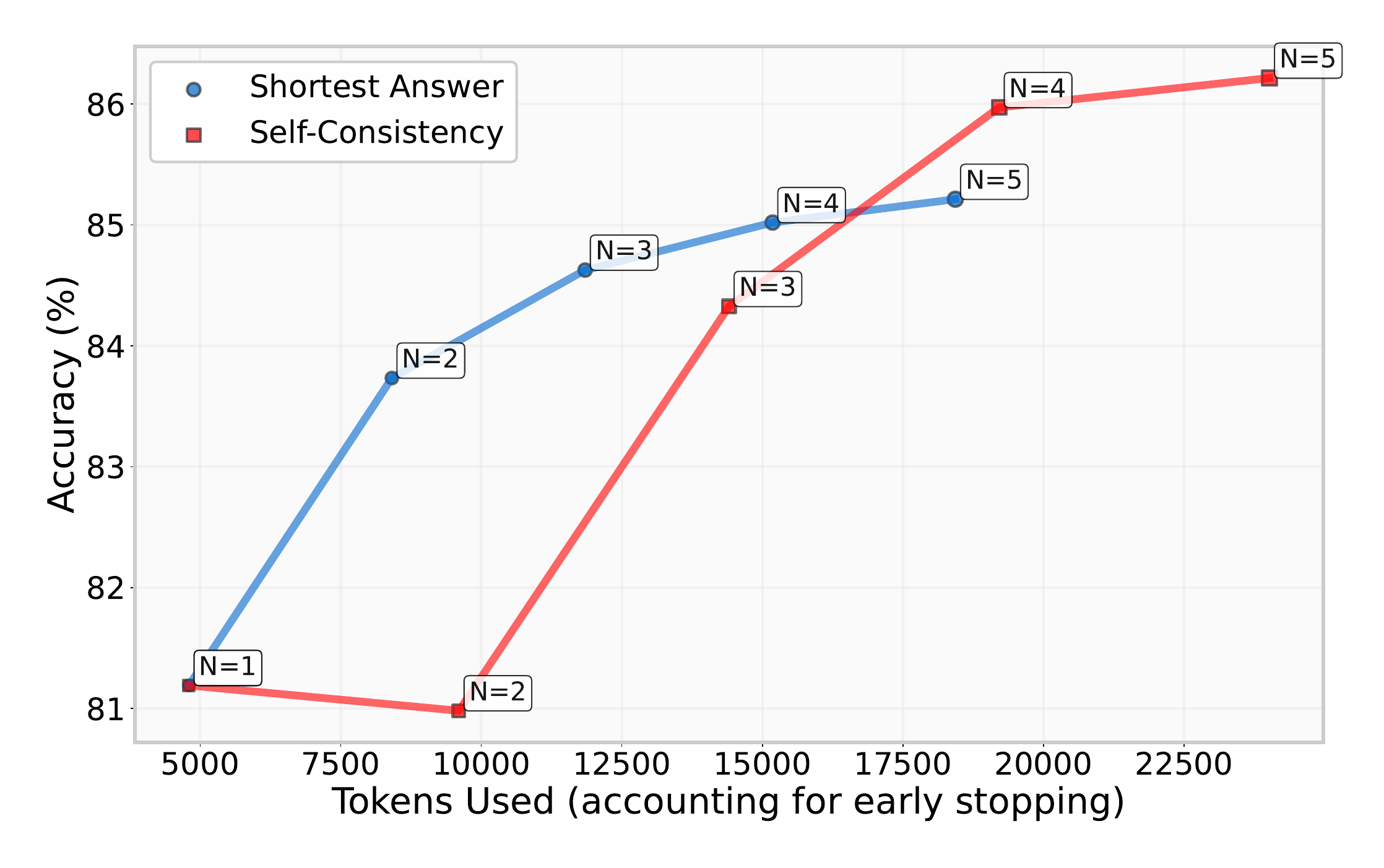}
\label{fig:pareto_grok3}
}
\hfill
\subfigure[Pareto curve for Qwen3-32B.]{
\includegraphics[width=0.48\textwidth]{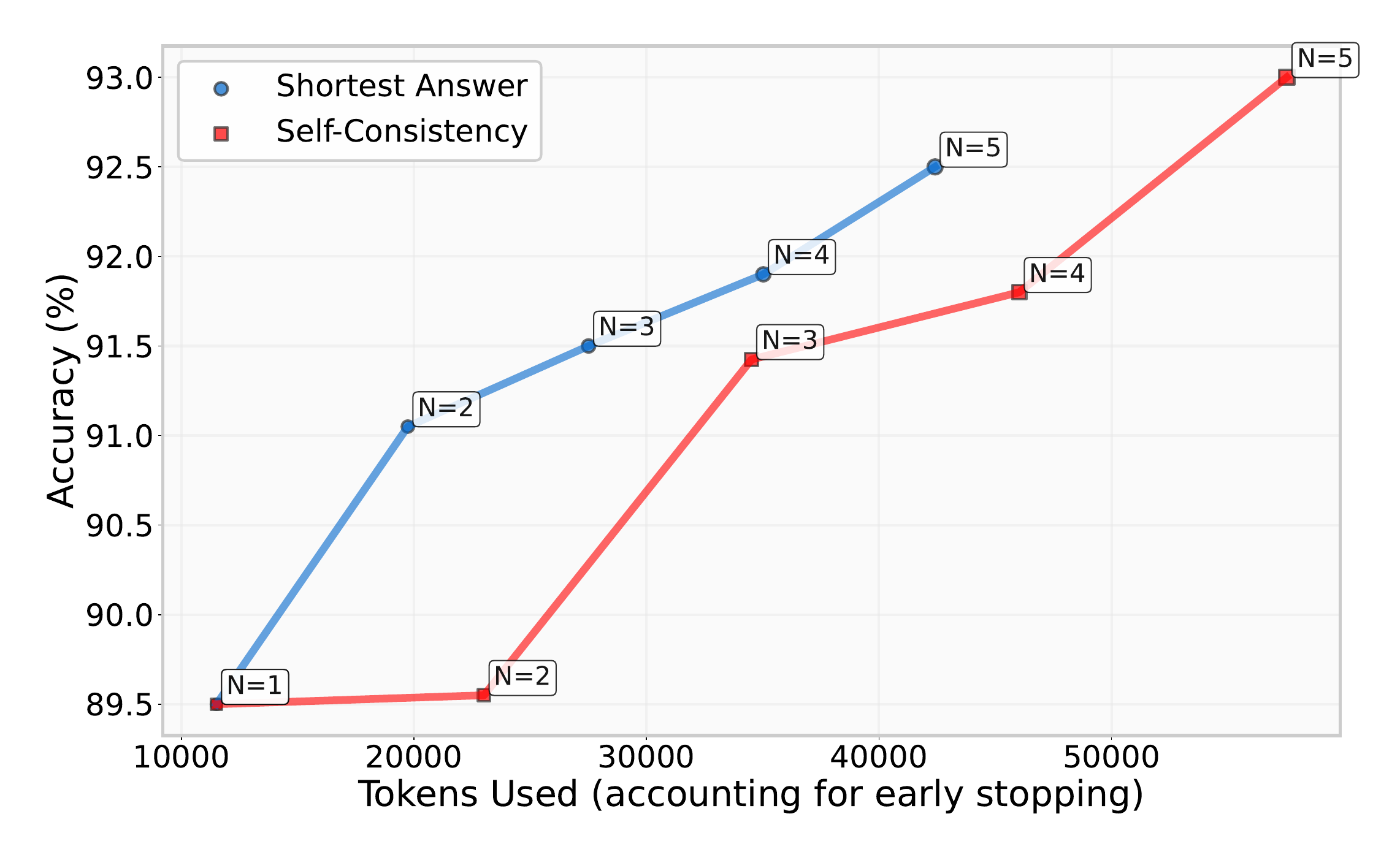}
\label{fig:pareto_Qwen}
}
\caption{Pareto curves comparing accuracy versus token usage for Grok-3-mini and Qwen3-32B on AIME.}
\label{fig:pareto_curves_combined}
\end{figure}

\subsection{Critical Point Analysis}
\label{app:critical_point}
\begin{figure}[htbp]
\centering
\subfigure[Token usage (AIME).]{
\includegraphics[width=0.48\textwidth]{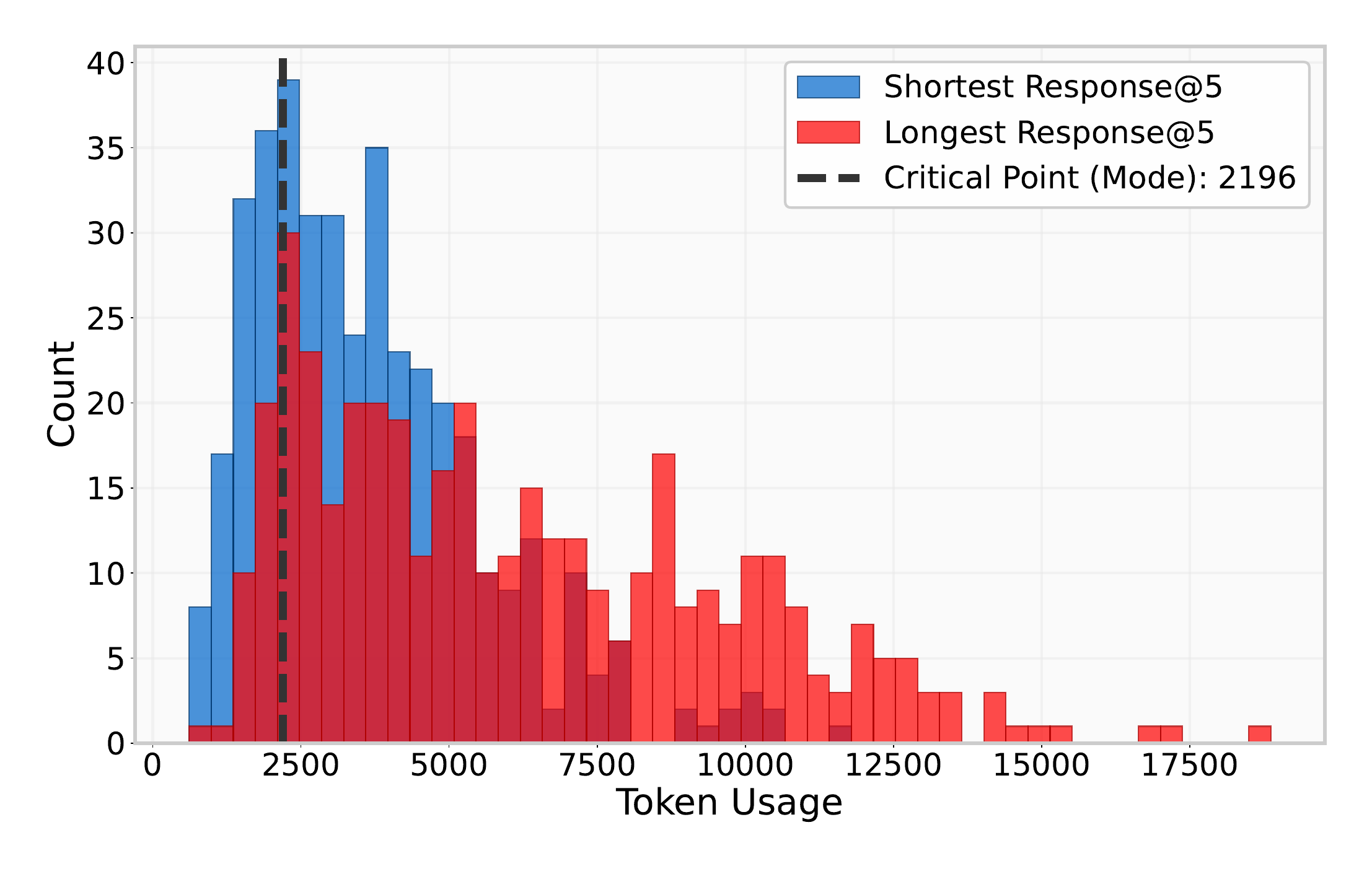}
\label{fig:grok_aime_token}
}
\hfill
\subfigure[Uncertainty (AIME).]{
\includegraphics[width=0.48\textwidth]{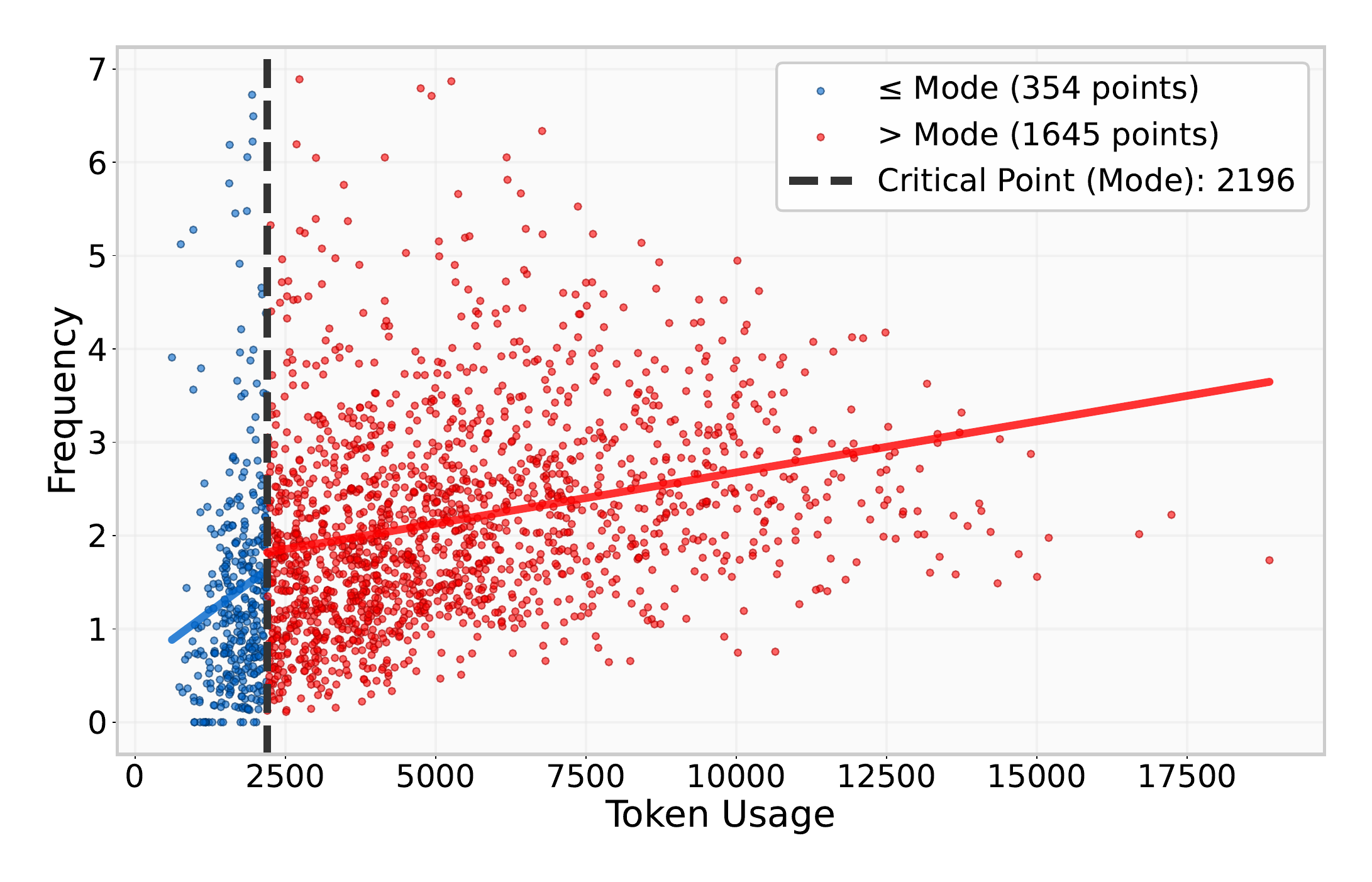}
\label{fig:grok_aime_uncertainty}
}
\par\medskip
\subfigure[Token usage (LiveCodeBench).]{
\includegraphics[width=0.48\textwidth]{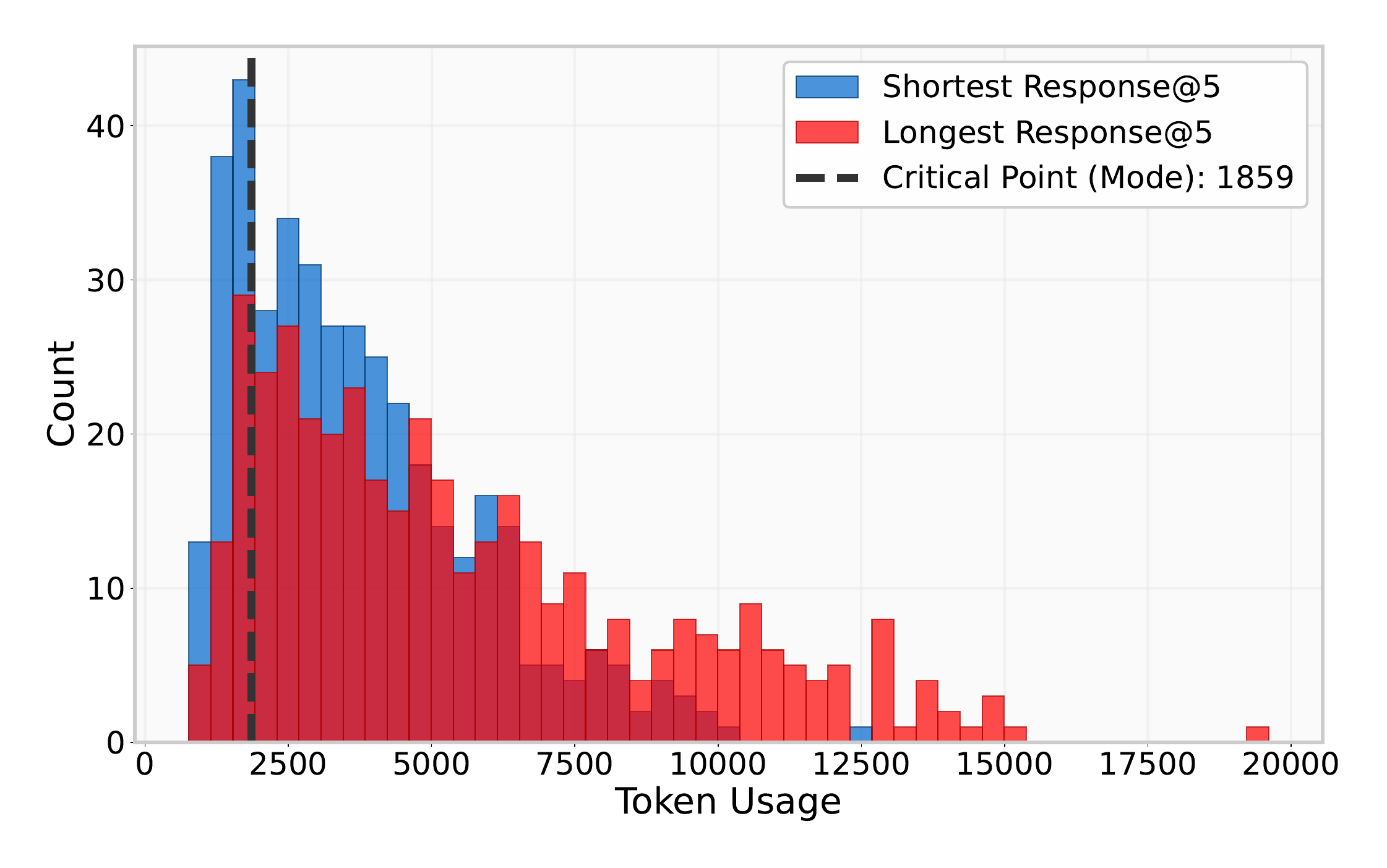}
\label{fig:grok_lcb_token}
}
\hfill
\subfigure[Uncertainty (LiveCodeBench).]{
\includegraphics[width=0.48\textwidth]{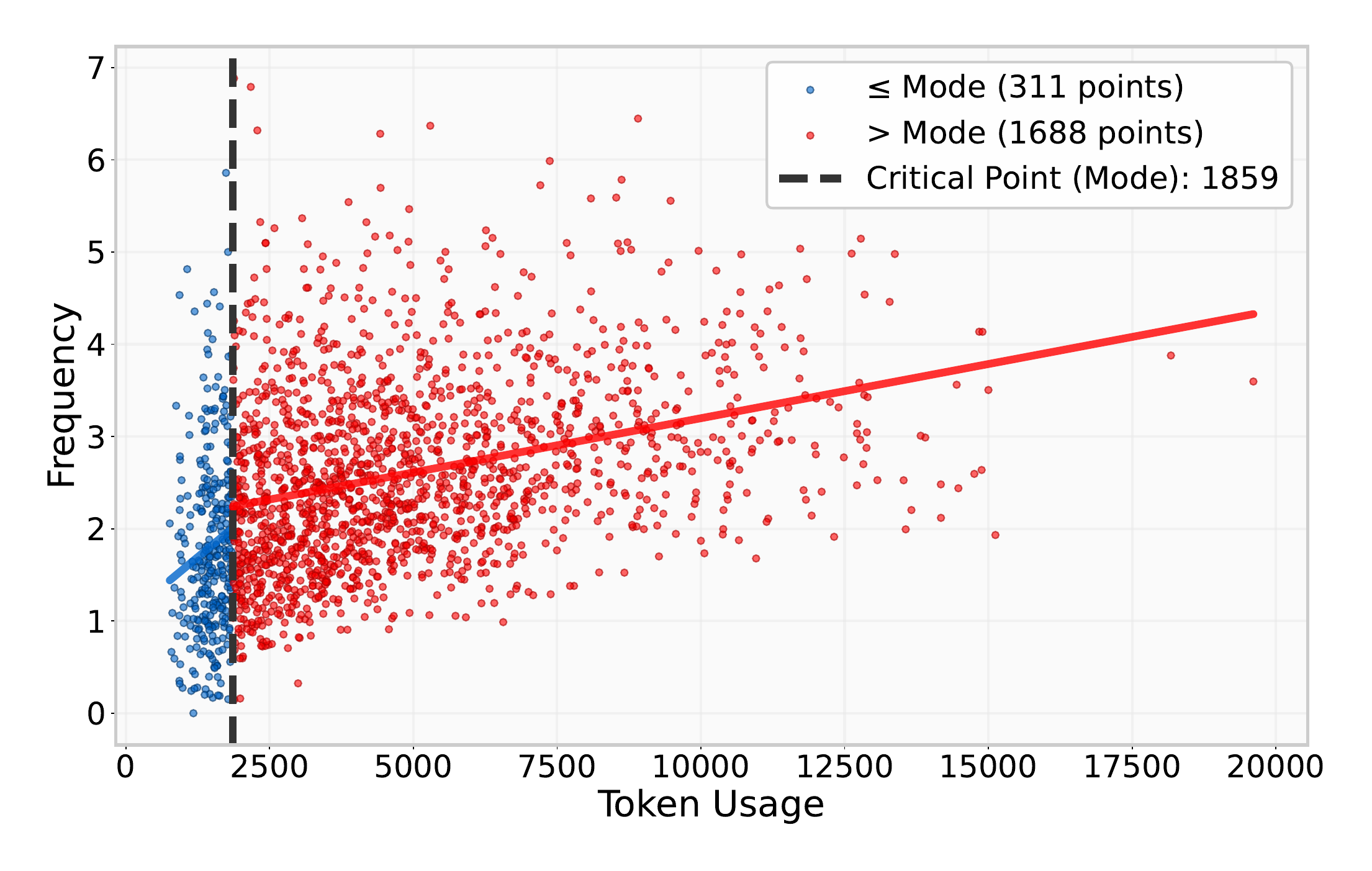}
\label{fig:grok_lcb_uncertainty}
}
\caption{Token-usage and critical point analyses for Grok-3-mini on AIME and LiveCodeBench.}
\label{fig:crit_grok}
\end{figure}

\FloatBarrier

\begin{figure}[htbp]
\centering
\subfigure[Token usage (AIME).]{
\includegraphics[width=0.48\textwidth]{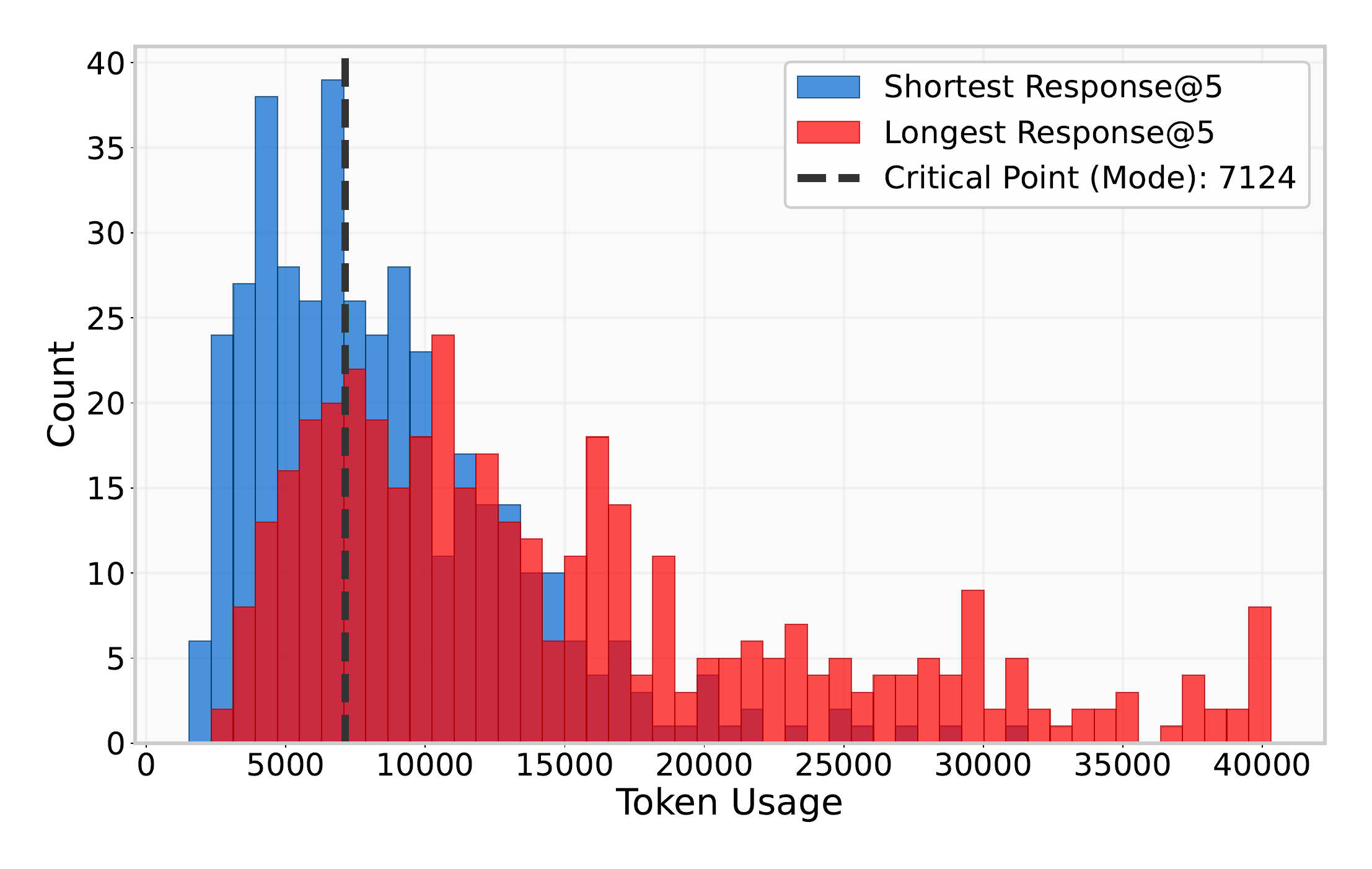}
\label{fig:qwen_aime_token}
}
\hfill
\subfigure[Uncertainty (AIME).]{
\includegraphics[width=0.48\textwidth]{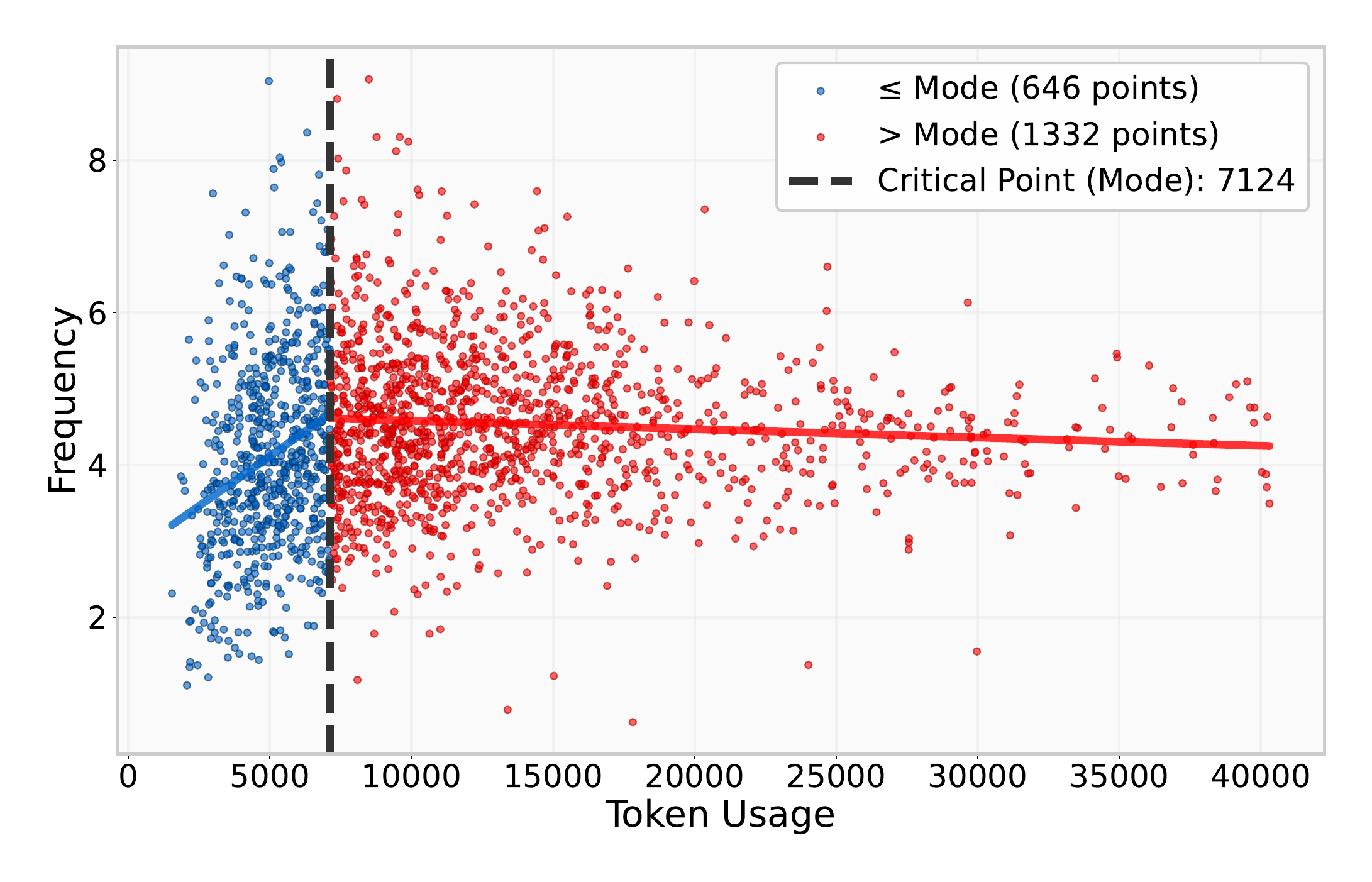}
\label{fig:qwen_aime_uncertainty}
}
\par\medskip
\subfigure[Token usage (LiveCodeBench).]{
\includegraphics[width=0.48\textwidth]{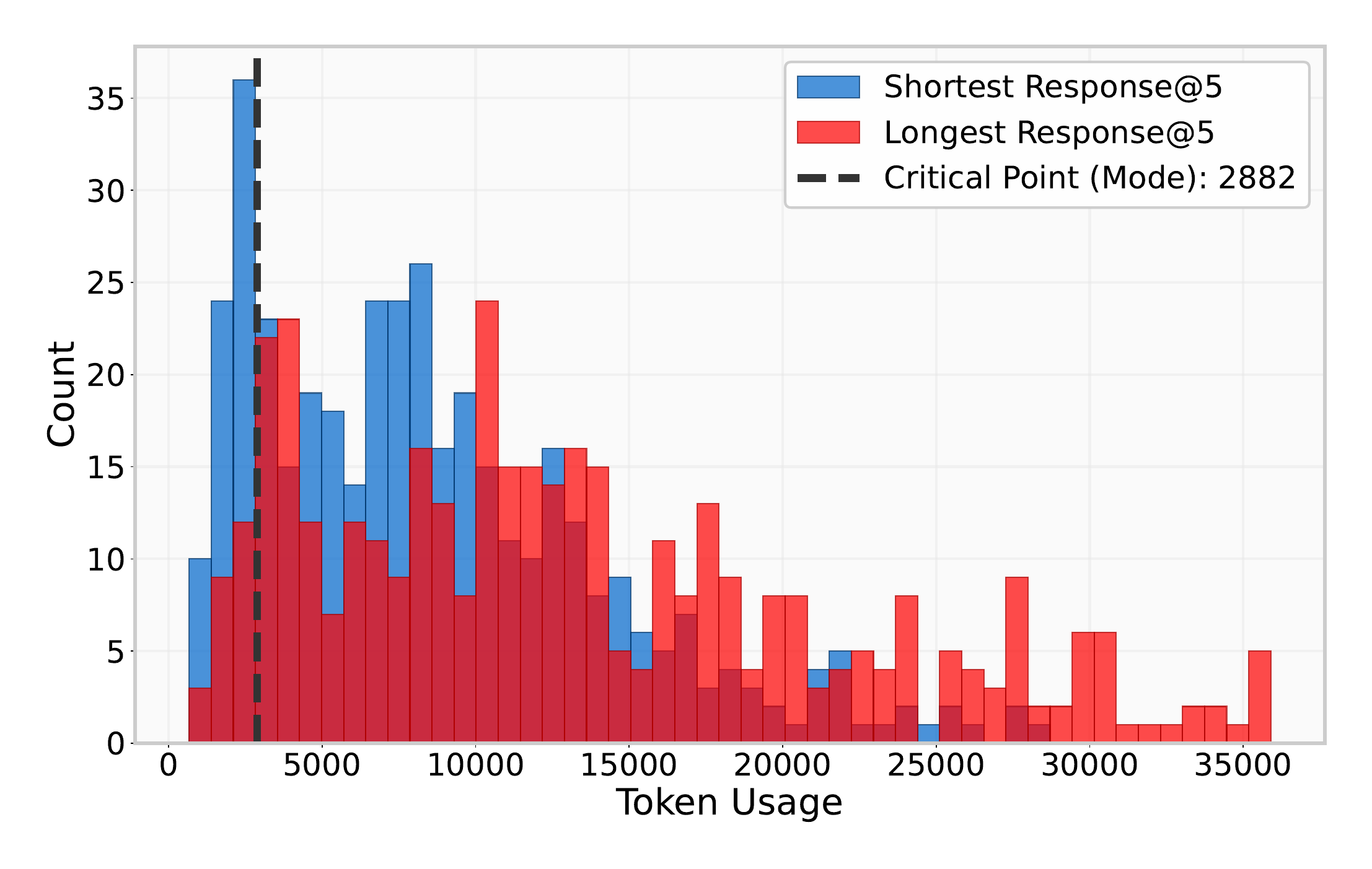}
\label{fig:qwen_lcb_token}
}
\hfill
\subfigure[Uncertainty (LiveCodeBench).]{
\includegraphics[width=0.48\textwidth]{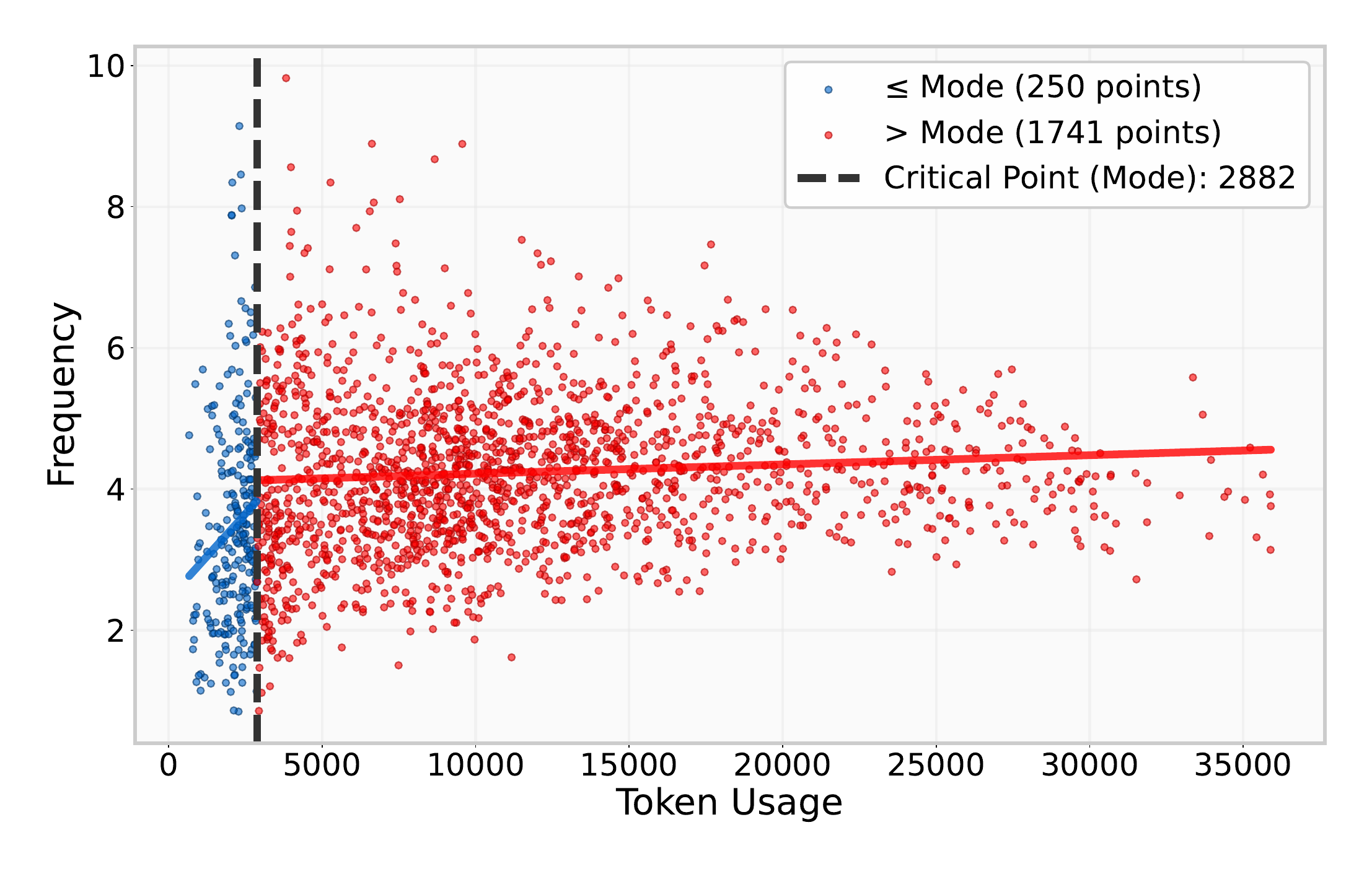}
\label{fig:qwen_lcb_uncertainty}
}
\caption{Token-usage and critical point analyses for Qwen3-32B on AIME and LiveCodeBench.}
\label{fig:crit_qwen}
\end{figure}

\FloatBarrier

\begin{figure}[htbp]
\centering
\subfigure[Token usage (LiveCodeBench).]{
\includegraphics[width=0.48\textwidth]{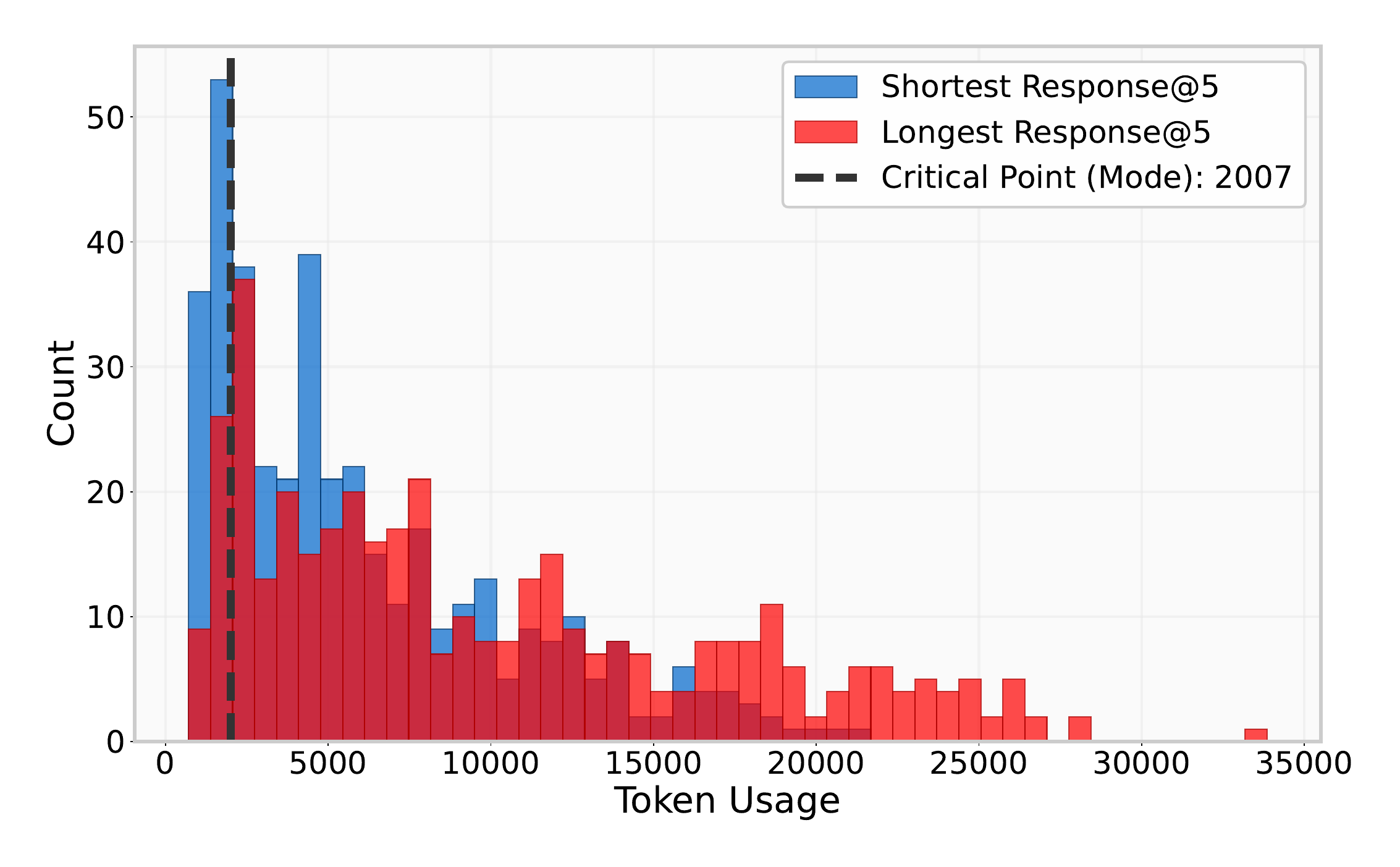}
\label{fig:deepseek_lcb_token}
}
\hfill
\subfigure[Uncertainty (LiveCodeBench).]{
\includegraphics[width=0.48\textwidth]{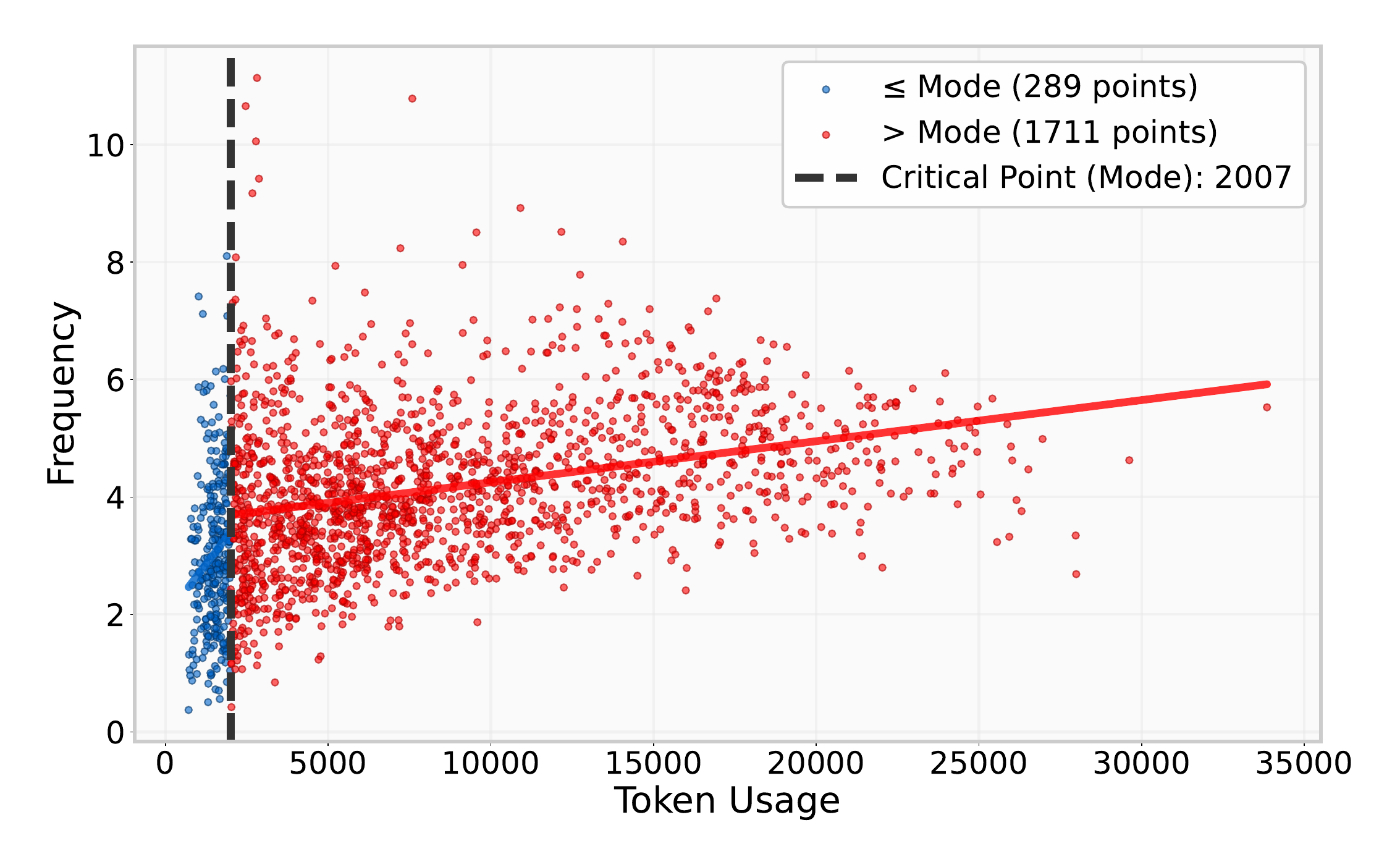}
\label{fig:deepseek_lcb_uncertainty}
}
\caption{Token-usage and critical point analyses for DeepSeek-R1 on LiveCodeBench. AIME results appear in the main paper.}
\label{fig:crit_deepseek}
\end{figure}

\FloatBarrier
\clearpage

\section{Prompts}
\label{app:prompts}
% Color definitions based on style guide
\definecolor{FigureBackground}{HTML}{FAFAFA}  % Very light gray for backgrounds
\definecolor{FigureSpines}{HTML}{CCCCCC}      % Medium gray for frames
\definecolor{PrimaryBlue}{HTML}{0066CC}       % Saturated blue
\definecolor{PrimaryOrange}{HTML}{FF8800}     % Saturated orange

% Academic-style prompt box
\newtcolorbox{promptbox}[2][FigureSpines]{
  enhanced,
  colback=FigureBackground,
  colframe=#1,
  fonttitle=\bfseries,
  fontupper=\ttfamily,
  boxrule=0.75pt,
  arc=0mm,                      % Sharp corners for academic look
  boxsep=3mm,
  left=5mm,
  right=5mm,
  top=3mm,
  bottom=3mm,
  title=#2,
  toptitle=2mm,
  bottomtitle=2mm,
  titlerule=0.5pt,              % Simple line under title
  colbacktitle=FigureBackground, % Same background as content
  coltitle=black,               % Black text for academic style
}

This section details the exact system prompts for the experiments.
\vspace{0.5cm}

\begin{promptbox}{AIME System Prompt}
Give a numerical answer from 0-999 in this format, \textbackslash{}boxed\{answer\}, so if the answer is 123, you should return \textbackslash{}boxed\{123\}.

[problem statement appended in user prompt]
\end{promptbox}

\vspace{0.5cm}

\begin{promptbox}{LiveCodeBench System Prompt}
You are an expert Python programmer. You will be given a question and will generate a correct Python program that matches the specification and passes all tests.

[problem statement and public tests appended in user prompt]
\end{promptbox}

\FloatBarrier
\section{List of Uncertainty Markers}
\label{app:uncertainty_markers}
The following list presents the uncertainty markers we use in our analysis to identify hedging, doubt, and self-correction patterns in model responses.

\begin{itemize}
    \item \texttt{maybe}, \texttt{alternatively}, \texttt{but}, \texttt{wait}, \texttt{perhaps}
    \item \texttt{possibly}, \texttt{probably}, \texttt{likely}, \texttt{might}, \texttt{may}, \texttt{could}, \texttt{would}
    \item \texttt{seems}, \texttt{appears}, \texttt{looks like}, \texttt{suggests}, \texttt{indicates}
    \item \texttt{i think}, \texttt{i believe}, \texttt{i guess}, \texttt{i suppose}, \texttt{i assume}
    \item \texttt{not sure}, \texttt{unclear}, \texttt{confusing}, \texttt{complicated}, \texttt{difficult}
    \item \texttt{however}, \texttt{although}, \texttt{though}, \texttt{unless}, \texttt{except}
    \item \texttt{on the other hand}, \texttt{actually}, \texttt{in fact}, \texttt{rather}, \texttt{hmm}
    \item \texttt{hold on}, \texttt{let me check}, \texttt{let me reconsider}
    \item \texttt{that's not right}, \texttt{i made an error}, \texttt{correction}, \texttt{mistake}, \texttt{check again}
    \item \texttt{if}, \texttt{assuming}, \texttt{suppose}, \texttt{what if}, \texttt{in case}, \texttt{provided that}
\end{itemize}

%%%%%%%%%%%%%%%%%%%%%%%%%%%%%%%%%%%%%%%%%%%%%%%%%%%%%%%%%%%%

\bibliographystyle{unsrtnat} % referencias em ordem textual

\end{document}